\documentclass[10pt, a4paper]{article}
\usepackage[final]{_external/lrec}

\usepackage{todonotes}

\usepackage{amsmath}
\usepackage{amssymb}

\usepackage{booktabs}
\usepackage{multirow}
\usepackage{makecell}
\usepackage{threeparttable}

\title{
    Next Reply Prediction X Dataset:\\Linguistic Discrepancies in Naively Generated Content 
}

 \name{
     Simon Münker$^1$, 
     Nils Schwager$^1$, 
     Kai Kugler$^1$,
     Michael Heseltine$^2$,
     Achim Rettinger$^1$
     \\
 } 

 \address{
     \begin{tabular}{c c}
         $^1$ Trier University, Computational Linguistics & 
         $^2$ University of Oxford, Sociology
         \\
         Universitätsring 15, 54296 Trier, Germany & 
         42-43 Park End Street, Oxford OX1 1JD, England 
         \\
         \{muenker, schwager, kuglerk, rettinger\}@uni-trier.de & michael.heseltine@sociology.ox.ac.uk \\
     \end{tabular}
     \\
 }

\abstract{
    The increasing use of Large Language Models (LLMs) as proxies for human participants in social science research presents a promising, yet methodologically risky, paradigm shift. While LLMs offer scalability and cost-efficiency, their "naive" application, where they are prompted to generate content without explicit behavioral constraints, introduces significant linguistic discrepancies that challenge the validity of research findings. This paper addresses these limitations by introducing a novel, history-conditioned reply prediction task on authentic $\mathbb{X}$ (formerly Twitter) data, to create a dataset designed to evaluate the linguistic output of LLMs against human-generated content. We analyze these discrepancies using stylistic and content-based metrics, providing a quantitative framework for researchers to assess the quality and authenticity of synthetic data. Our findings highlight the need for more sophisticated prompting techniques and specialized datasets to ensure that LLM-generated content accurately reflects the complex linguistic patterns of human communication, thereby improving the validity of computational social science studies.
    \\\newline 
}

\begin{document}

\maketitleabstract

\section{Introduction}
\label{sec:introduction}

The widespread adoption of Large Language Models (LLMs) began with the release of ChatGPT and similar conversational AI systems, fundamentally transforming how humans interact with artificial intelligence \cite{aimeur2023fake}. This technological advancement created a paradigm shift in computational social science research, with LLMs increasingly positioned as viable proxies for human participants in behavioral studies \cite{park2023generative, perez2023serious}. The promise is compelling: LLMs offer unprecedented scalability, cost-efficiency, and the ability to conduct large-scale behavioral research without the traditional constraints of human participant recruitment, retention, and ethical complexities.

However, this anthropomorphic perspective introduces significant methodological risks, particularly when researchers employ naive applications that rely exclusively on prompt engineering without adequate consideration of underlying model limitations, training biases, and domain-specific validation requirements \cite{larooij2025large}. The quality and representativeness of training datasets become critically important as grounding mechanisms, especially for socially sensitive tasks where cultural nuance, contextual understanding, and authentic human judgment remain central to meaningful analysis. While earlier concerns focused on detecting artificial or malicious content, contemporary LLMs produce increasingly sophisticated outputs that superficially mimic human communication patterns \cite{crothers2023machine}. This evolution makes validation more critical yet paradoxically more challenging: the better LLMs become at generating plausible content, the more crucial it becomes to understand where and how they diverge from authentic human behavior.

Our paper addresses a fundamental question at the intersection of natural language processing and computational social science: \textit{Can current LLMs reliably replicate authentic human social media behavior patterns when tasked with user modeling applications?} This question becomes particularly pressing given the growing reliance on synthetic data in computational social science \cite{burgard2017synthetic}, where the assumption of authentic human-like generation underpins the validity of research findings. We approach this question through a systematic comparison between genuine X content and synthetic posts generated through both prompt-based and fine-tuned approaches, examining linguistic discrepancies across multiple analytical dimensions.

\subsection{Research Questions and Hypotheses}
We investigate three primary research questions using a self-collected German and English X dataset:

\begin{description}
    \item[RQ1] 
    To what extent do LLM-generated social media posts exhibit detectable linguistic patterns that distinguish them from authentic human content across quantitative, morphological, and semantic dimensions?

    \item[RQ2]
    How does fine-tuning on domain-specific social media data improve the linguistic authenticity of generated content compared to prompt-based generation approaches?

    \item[RQ3]
    Can machine learning classifiers reliably distinguish between human and synthetic social media content, and what features prove most discriminative?
\end{description}

Building on empirical evidence from related work on LLM limitations in social simulation \cite{liu2022quantifying, hershcovich2022challenges, munker2025don}, we hypothesize that while LLMs can produce individually plausible social media posts, systematic analysis reveals consistent linguistic signatures that enable reliable detection of synthetic content. Furthermore, we anticipate that fine-tuned models will show reduced but still detectable deviation patterns compared to prompt-based approaches. Related research confirms that fine-tuned models outperform prompt-based approaches in social simulations \cite{lin2024designing} and text annotation tasks \cite{alizadeh2025open} in human-LLM alignment. 

\subsection{Our Contributions}
Our work makes three primary contributions to the language resources and evaluation community:

\begin{enumerate}
    \item 
    We publish a history-conditioned reply prediction dataset for X content, comprising authentic human posts alongside corresponding synthetic generations using both prompt-based and fine-tuned approaches across English and German languages. (Sec. \ref{sec:methods:dataset})

    \item 
    We present a multi-dimensional evaluation framework combining multiple layers of quantitative linguistics analysis to assess human-machine linguistic alignment. (Sec. \ref{sec:methods:evaluation})

    \item 
    We conduct a comparison of encoder (tf-idf, static dense, transformer) and feature (see above) combinations for detecting the synthetic content. (Sec. \ref{sec:methods:task})
\end{enumerate}
\section{Background}
\label{sec:background}

\subsection{LLMs as Human Simulacra}
The emergence of Large Language Models has fundamentally transformed computational social science research, with contemporary studies increasingly positioning LLMs as human simulacra \cite{park2023generative} capable of simulating complex user behaviors through sophisticated text-to-text engagement \cite{larooij2025large, munker2025don}. This paradigm shift offers compelling advantages including cost reduction, ethical compliance, and enhanced scalability for large-scale behavioral studies \cite{perez2023serious, thapa2025large}.

However, empirical validation reveals significant limitations in the authenticity of LLM-generated social behavior. Studies demonstrate systematic biases in the diversity of political \cite{liu2022quantifying, munker2025political} and cultural \cite{hershcovich2022challenges} positions represented in current LLMs. These limitations challenge the prevalent assumption that LLMs can serve as reliable human proxies, particularly when researchers employ naive applications that rely exclusively on prompt engineering without adequate consideration of underlying model limitations, training biases, and domain-specific validation requirements.

The anthropomorphic perspective introduces methodological risks that become especially problematic in applications requiring nuanced social understanding. While individual LLM-generated texts may appear plausible, systematic analysis often reveals consistent linguistic signatures that distinguish synthetic from authentic content. This detectability gap has important implications for the ecological validity of LLM-based simulations in social research contexts, where the assumption of authentic human-like generation underpins the validity of research findings.

\subsection{Synthetic Content Detection in Social Media}
The field of synthetic content detection has evolved significantly alongside advances in generation capabilities. Traditional approaches to misinformation detection on social media platforms target artificial or malicious content from regular users \cite{yang2019unsupervised} and develop comprehensive bot detection systems \cite{hayawi2023social}. However, the current generation of LLMs produces increasingly sophisticated outputs that closely mimic human communication patterns, making detection more challenging and validation more critical.

Recent advances in AI-generated content detection \cite{chong2023bot, abburi2024toward} reveal that even sophisticated generation techniques exhibit systematic linguistic patterns across multiple dimensions. These patterns manifest in quantitative features (complexity, readability, lexical diversity), morphosyntactic structures (part-of-speech distributions, syntactic complexity), and semantic distributions (topic diversity, emotion patterns, sentiment biases). The persistence of these linguistic signatures across different generation approaches suggests fundamental limitations in current language modeling techniques for authentic social media simulation.

Our work motivates a shift toward multi-dimensional evaluation frameworks that capture the full spectrum of linguistic differences between human and synthetic content. Surface-level plausibility assessments prove insufficient for validating LLM-generated social media content, necessitating comprehensive protocols that examine linguistic authenticity across quantitative, morphological, and semantic dimensions simultaneously. We build upon these methodological foundations while introducing a novel history-conditioned dataset and systematic comparison of detection approaches, addressing the critical gap between generation capability and authentic behavioral replication.
\section{Methods}
\label{sec:methods}

\subsection{Data: Authentic vs. Synthetic}
\label{sec:methods:dataset}

\paragraph{Collection/Preprocessing}
Our final dataset is based on two raw data dumps – English and German – collected from X. The sets are collected around keywords concerning the political discourses in the US and Germany during the first half of 2023. The samples contain two types of content: a) Tweets (posts) and (b) replies from X users towards these tweets (DE: $3.381.111$, EN: $7.790.741$).

First, we group all first-order replies with the tweets to which they are responding, creating tweet-reply pairs that preserve conversational context. We then reorganize these samples by user, resulting in subsets containing each user's complete reply history along with the original tweets they responded to.

Next, we apply two preprocessing steps to ensure data quality. First, we remove tweet-reply pairs containing URLs (images, GIFs, and links), as these cannot be properly processed by the LLM and the classifiers. Second, we remove users with the highest reply frequencies (DE: $5\%$, EN: $1\%$; Quotas result in max DE: $24$, EN: $21$ samples per user) and split the remaining users into train and test sets. This ensures our analysis captures the model's ability to learn generalizable linguistic styles across the user population, rather than memorizing patterns of individual users.

\paragraph{Transformation}
We construct a History-Conditioned Reply Prediction Task \cite{munker2025don}, using the native instruction-completion format of instruction-LLMs: three tweet-reply pairs as "history", along with a fourth tweet for the model to respond to. We add a system prompt: \textit{You are a social media user responding to conversations. Keep your replies consistent with your previous writing style and the perspectives you have expressed earlier.} This conditions the LLM by presenting the tweet-reply history as if it had already generated those replies during prior turns. 

This approach offers three advantages: (1) the model learns from authentic behavioral patterns without hand-crafted features encoding response characteristics; (2) It allows synthetic sample generation without further training only by prompting (3) during fine-tuning, the withheld fourth reply serves as the supervised target.

\paragraph{Fine-Tuning} 
We fine-tune Qwen3 8B \cite{yang2025qwen3} for each language variant using supervised learning with loss computed exclusively on the last assistant responses. Both training datasets are sub-sampled to $5000$ examples. Training uses a warm-up ratio of $0.1$ and single-epoch optimization with otherwise default hyperparameter (e.g.; learning rate of $2\mathrm{e}{-}5$). Each model trains for approximately 100 minutes on an NVIDIA L40S GPU with 48GB of VRAM.

\paragraph{Generation}
We generate a single synthetic reply per test prompt using both the base and fine-tuned Qwen3 8B models. Generation uses Qwen3's default sampling parameters (\texttt{temperature}: $0.6$, \texttt{top\_k}: $20$, \texttt{top\_p}: $0.9$) with a maximum output length of $200$ tokens. No post-generation filtering is applied. All model outputs are retained regardless of length, coherence, or formatting. 

\paragraph{Published Dataset}
The published dataset \textit{(GitHub repository, see Sec.~\ref{sec:methods:code})} serves as the foundation for all subsequent analyses. It consists of $1000$ samples per language, each containing: \texttt{prompt} (three historical tweet-reply pairs plus fourth tweet in chat completion format), \texttt{authentic reply} (ground truth from test users), and two generated columns \texttt{base model reply} and \texttt{ft model reply} produced by applying the generation procedure described above to the base and fine-tuned Qwen3 8B models respectively. As a scientific artifact, the dataset serves three potential usages: 1) improving the next reply prediction task given our proposed metrics, 2) developing additional metrics to analyze the LLM-human alignment further, and 3) improving synthetic content detection classifiers.

\subsection{Evaluation: Levels of Alignment}
\label{sec:methods:evaluation}

\paragraph{Quantitative Features} 
We implement the complete NeLa feature suite \cite{horne2018assessing} through a modular extraction pipeline spanning five linguistic dimensions: complexity, style, bias, affect, and moral reasoning patterns. The system extracts linguistic profiles including type-token ratios, average sentence length, lexical diversity measures, readability scores (Flesch-Kincaid \cite{kincaid1975derivation}, Gunning Fog \cite{gunning1952technique}), and character-level complexity metrics.

\paragraph{Morphosyntactic Extraction}
Using the spaCy processing pipeline \cite{montani2023spacy}, we extract comprehensive linguistic annotations including part-of-speech tag distributions following Universal Dependencies standards \cite{de2021universal}, named entity recognition patterns across 18 standard categories (PERSON, ORG, GPE, DATE, etc.), dependency relation frequencies, and syntactic complexity measures. Our implementation computes frequency-normalized distributions for both POS categories and NER labels, incorporating lexical diversity metrics and average sentence length measurements.

\paragraph{Semantic Classification}
We employ the TweetEval benchmark \cite{barbieri2020tweeteval} through pre-trained transformer-based classifiers to evaluate content across multiple semantic dimensions. Our pipeline integrates three specialized models: \texttt{tweet-topic-21-multi} \cite{antypas-etal-2022-twitter} for topic classification, \texttt{twitter-RoBERTa-base-emotion} \cite{camacho-collados-etal-2022-tweetnlp} for emotion detection, and \texttt{twitter-RoBERTa-base-sentiment} \cite{barbieri2020tweeteval} for sentiment analysis.

\paragraph{Cluster-based Similarity}
Utilizing the state-of-the-art instruction-following embedding model Qwen3 \cite{zhang2025qwen3}, we compute semantic similarity distributions within and across content categories. Through cluster analysis using PCA dimensionality reduction \cite{pearson1901liii} and Affinity Propagation \cite{frey2007clustering}, we analyze the proportion of clusters per content category.

\paragraph{Feature-Vector Distance Computation}
To quantify linguistic alignment between human and synthetic content, we implement a distance-based similarity metric. For each corpus $C$ and linguistic feature set $F$, we compute normalized feature vectors through the following procedure:

\begin{enumerate}
    \item 
    Extract mean feature scores for each corpus-feature combination:
    $$\bar{f}_C^i = \frac{1}{|C|} \sum_{d \in C} F_i(d)$$
    where $d$ represents sample in corpus $C$ and $F_i$ denotes the $i$-th feature in $F$.

    \item 
    Construct corpus vectors: $\mathbf{v}_C = [\bar{f}_C^1, \ldots, \bar{f}_C^{|F|}]$

    \item Compute pairwise Cosine similarity between two corpus vectors defined as $s(\mathbf{v}_{C_1}, \mathbf{v}_{C_2})$.
\end{enumerate}

\subsection{Validation: Detecting Synthetics}
\label{sec:methods:task}

As a downstream validation task, we implement a comparison of detection approaches spanning the spectrum from traditional sparse representations to modern dense embeddings. We concatenate the above-described features with the following text embeddings to investigate if these features improve the identification of synthetically generated examples.

\paragraph{Encoding Approaches}

\begin{description}
    \item[TF-IDF]
    Sparse term frequency-inverse document frequency vectorization \cite{ramos2003using} with uni-gram features and lowercase normalization for, baseline, traditional bag-of-words representation.

    \item[FastText] 
    Dense 300-dimensional word vectors \cite{joulin2017bag} using spaCy's \texttt{en\_core\_web\_lg} and \texttt{de\_core\_news\_lg} model, aggregated through mean pooling for efficient semantic representation.

    \item[Qwen3 Embedding]: 
    State-of-the-art instruction-following embeddings using the Qwen/Qwen3-Embedding-8B model \cite{zhang2025qwen3} with a specialized authorship detection prompt: "Instruct: Find tweets with similar authorship patterns (human vs. AI-generated) based on writing style, vocabulary choice, and content structure". We choose Qwen3 as it shows benchmark-leading performance compared to the number of parameters in text classification tasks \cite{pan2025qwen3, heseltine2025comparing}.
\end{description}

\paragraph{Feature Combination Strategy}
To investigate the complementary nature of different representation types, we systematically evaluate all possible combinations of encoding approaches and extracted features, creating hybrid representations that capture multiple linguistic perspectives simultaneously.

\paragraph{Classification Model}
We utilize XGBoost (eXtreme Gradient Boosting) \cite{chen2016xgboost} as our classification algorithm. We select XGBoost for its promising performance on heterogeneous feature combinations, robust handling of different feature scales, and interpretability through feature importance analysis.

\subsection{Reproducibility and Code Availability}
\label{sec:methods:code}
All experimental procedures, statistical analyses, and model training protocols are implemented using open-source tools including scikit-learn \cite{sklearn_api}, spaCy \cite{montani2023spacy}, Transformers Reinforcement Learning (TRL) \cite{vonwerra2022trl} and Sentence Transformers \cite{reimers2019sentence}. 


\section{Results}
\label{sec:results}

Our results reveal systematic linguistic differences between human and synthetic content across all analytical dimensions, with fine-tuned models consistently showing superior alignment to human content compared to prompt-based approaches. These findings address our three research questions through complementary lenses: similarity analysis (RQ1 and RQ2) and classification performance (RQ3).

\subsection{Quantitative Linguistics Analysis}
Table \ref{tab.feature.similarity} presents the calculated similarity scores between corpus subsets across all feature extraction approaches. The results show a consistent hierarchy of alignment, with fine-tuned models ($F$) showing highest similarity to human original content ($O$), followed by moderate alignment between original and prompt-based content ($P$), while prompt-based and fine-tuned models exhibit the lowest mutual similarity.

\begin{table}[ht]
	\centering
	\begin{tabular}{l|r|r|r}
		\toprule
		{\small Feat./Lang.} & $s(O, P)$ & $s(O, F)$ & $s(P, F)$ \\
		\midrule
		\multicolumn{4}{c}{\textbf{Quantitative Features (NeLa)}} \\
		\midrule
		German      & 0.7908     & 0.8048     & 0.6995     \\
		English     & 0.8408     & 0.8957     & 0.8410     \\
		\midrule
		\multicolumn{4}{c}{\textbf{Morphosyntactic Extraction (SpaCy)}} \\
		\midrule
		German      & 0.9498     & 0.9748     & 0.9437     \\
		English     & 0.9423     & 0.9816     & 0.9357     \\
		\midrule
		\multicolumn{4}{c}{\textbf{Semantic Classification (TweetEval)}} \\
		\midrule
		German      & 0.9695     & 0.9874     & 0.9832     \\
		English     & 0.9745     & 0.9819     & 0.9786     \\
		\midrule
		\multicolumn{4}{c}{\textbf{Cluster-based Similarity}} \\
		\midrule
		German      & 0.8016     & 0.9713     & 0.7435     \\
		English     & 0.8977     & 0.9620     & 0.8323     \\
		\midrule
	\end{tabular}
	\caption{
		Comparison of the calculated similarity $s$ between the corpora subsets human original ($O$), synthetic only prompted ($P$) and synthetic fine-tuned ($F$) across German and English on all features described in section \ref{sec:methods:evaluation}. A higher value indicates a more aligned model behavior.
	}
	\label{tab.feature.similarity}
\end{table}

\paragraph{Quantitative Features}
The NeLa features reveal substantial differences in linguistic complexity and style patterns. For German, similarity between original and fine-tuned content reaches $0.8048$, significantly higher than the $0.6995$ similarity between prompt-based \& fine-tuned approaches. English demonstrates even stronger alignment patterns, with original \& fine-tuned similarity achieving $0.8957$, while original-prompt similarity reaches $0.8408$.

\begin{figure}[ht]
    \centering
    \includegraphics[width=1.0\linewidth]{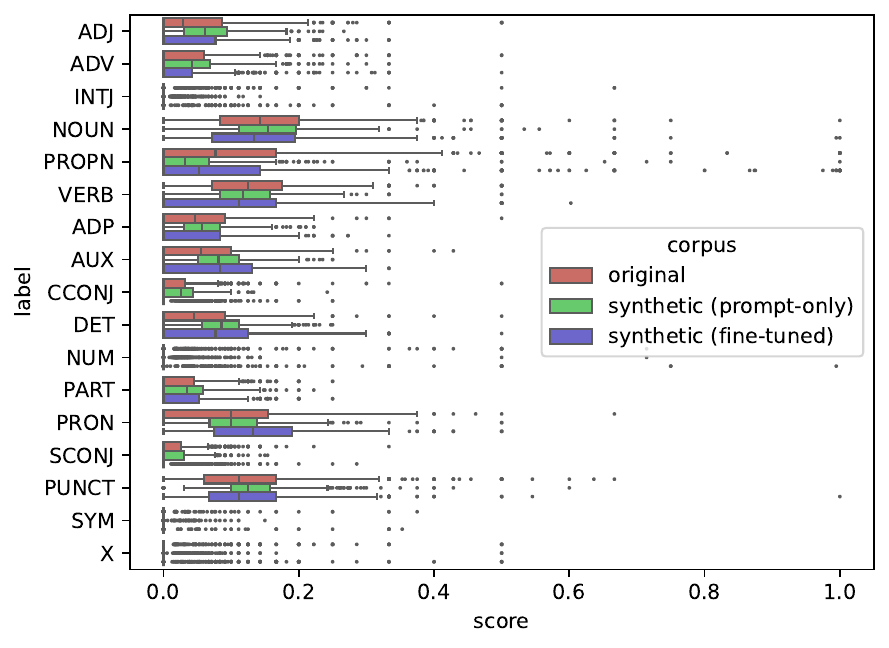}
    \caption{
        Locality, spread and skewness (x-axes) of each POS category (y-axes) for the English corpus split into subsets.
    }
    \label{fig.feature.spacy.pos.english}
\end{figure}

\paragraph{Morphosyntactic Extraction}
The Morphosyntactic analysis reveals the highest overall similarity scores across all approaches. German shows a high alignment between original \& fine-tuned content ($0.9748$), with prompt-based models achieving $0.9498$ similarity to original content. However, detailed examination reveals that prompt-based models exhibit distinctive usage patterns, particularly in coordinating (CCONJ) and subordinating conjunctions (SCONJ) (Figure \ref{fig.feature.spacy.pos.english}).

\begin{figure*}[ht]
    \centering
    \includegraphics[width=1.0\linewidth]{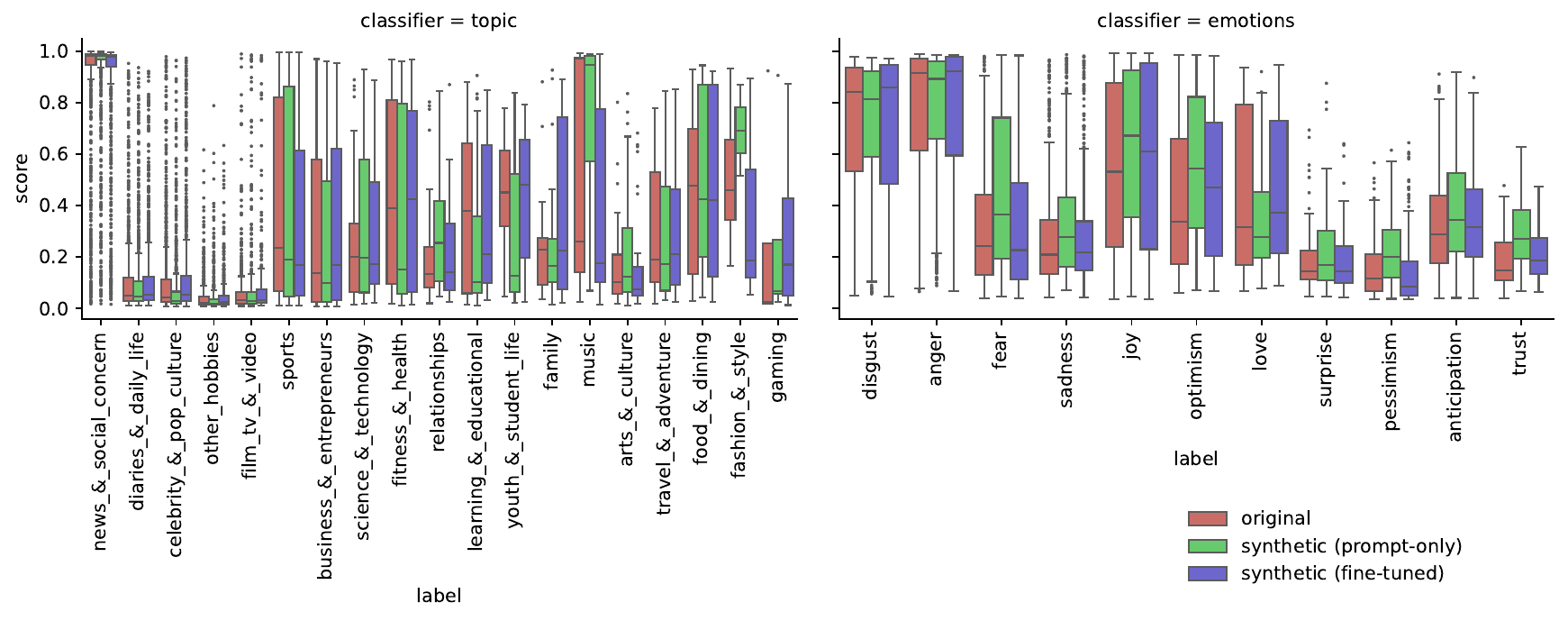}
    \caption{
        Locality, spread and skewness (y-axes) of the TweetEval topic and emotion classifier (x-axes) for the English corpus split into subsets.
    }
    \label{fig.tweet_eval.box.topic_emotions.english}
\end{figure*}

\paragraph{Semantic Classification}
Semantic classification shows the most consistent alignment across all model types, with similarity scores exceeding $0.97$ in all comparisons. German achieves the highest alignment between prompt and fine-tuned models ($0.9832$), while English shows marginally lower but still substantial similarity ($0.9786$). Despite these high similarity scores, qualitative analysis reveals that prompt-based models generate more topically diverse content and exhibit significantly higher proportions of positive emotion classifications compared to human content (Figure \ref{fig.tweet_eval.box.topic_emotions.english}).

\paragraph{Cluster-based Similarity}
Embedding-based cluster analysis reveals the most pronounced differences between generation approaches. Fine-tuned models achieve high alignment with original content (German: $0.9713$, English: $0.9620$), while prompt-based models show notably lower similarity to both original content and fine-tuned variants. The substantial gap between original \& prompt similarity (German: $0.8016$, English: $0.8977$) and original \& fine-tuned similarity demonstrates that semantic distributional properties are particularly sensitive.

\subsection{Validation Task}
Table \ref{tab.task.prediction} presents the classification results for distinguishing between human original ($O$), synthetic fine-tuned ($F$), and synthetic prompted ($P$) content across various feature combinations and encoding approaches. The results consistently demonstrate that prompt-based synthetic content ($P$) achieves the highest detection accuracy, while fine-tuned content ($F$) proves most challenging to distinguish from human original content ($O$).

\begin{table}[ht]
    \centering
    \begin{tabular}{l|r|r|r|r}
         \toprule
		{\small Feat./Lang.} & $F1(O)$ & $F1(F)$ & $F1(P)$ & avg \\
        \midrule
        \multicolumn{5}{c}{\textbf{tf–idf + fastText + \{TweetEval, SpaCy, NeLa\}}} \\
        \midrule
		German      & 0.6666     & 0.6938     & 0.8297 & 0.7301     \\
		English     & 0.6534     & 0.6382     & 0.8000 & 0.6972    \\
        \midrule
        \multicolumn{5}{c}{\textbf{tf–idf + fastText + Qwen}} \\
        \midrule
		German      & 0.6336     & 0.6476     & 0.8510 & 0.7107     \\
		English     & 0.5531     & 0.6136     & 0.6725 & 0.6240    \\
		\midrule
        \multicolumn{5}{c}{\textbf{Qwen}} \\
        \midrule
		German      & 0.5800     & 0.5849     & 0.7446 & 0.6365     \\
		English     & 0.5544     & 0.5940     & 0.8163 & 0.6549    \\
        \midrule
        \multicolumn{5}{c}{\textbf{fastText}} \\
        \midrule
		German      & 0.6734     & 0.6407     & 0.7878 & 0.7007    \\
		English     & 0.5858     & 0.6136     & 0.6725 & 0.6240    \\
        \midrule
        \multicolumn{5}{c}{\textbf{tf–idf}} \\
        \midrule
		German      & 0.6400     & 0.5961     & 0.8333 & 0.6898     \\
		English     & 0.5200     & 0.5000     & 0.7800 & 0.6000    \\
		\bottomrule
    \end{tabular}
    \caption{
        Results of the detection task for the individual $F1$ scores per class, human original ($O$), synthetic only prompted ($P$) and synthetic fine-tuned ($F$), and the macro average across German and English on a selected range of feature combinations with XGBoost as classifier.
    }
    \label{tab.task.prediction}
\end{table}

\paragraph{Feature Combination Performance} 
The most effective approach combines tf-idf, fastText embeddings, and one or more extracted features (TweetEval, SpaCy, NeLa), achieving macro $F1$ scores of $0.7301$ for German and $0.6972$ for English. Notably, prompt-based content consistently achieves the highest individual $F1$ scores across all feature combinations (German: $0.8297-0.8510$, English: $0.6725-0.8163$).

\paragraph{Encoding Approach Analysis}
Modern embedding approaches show competitive but not superior performance compared to traditional methods. The Qwen embedding model alone achieves moderate performance (German $F1$: $0.6365$, English $F1$: $0.6549$), while fastText embeddings demonstrate strong baseline performance (German $F1$: $0.7007$, English $F1$: $0.6240$). Surprisingly, simple tf-idf representations prove remarkably effective, particularly for prompt-based content detection (German: $0.8333$, English: $0.7800$).
\section{Discussion}
\label{sec:discussion}

\subsection{Implications for Computational Social Science}
Our findings reveal fundamental challenges for the ecological validity of LLM-based simulations in social research contexts. While current generation techniques can produce individually plausible social media posts, systematic analysis reveals consistent patterns that distinguish synthetic from authentic content across multiple linguistic dimensions. The detection accuracies achieved in our validation task, particularly for prompt-based content, indicate persistent linguistic signatures that compromise the authenticity of LLM-generated social media discourse.

This detectability gap has important implications for applications in computational social science, where researchers increasingly rely on LLMs as human proxies for behavioral studies. The systematic differences we observe in quantitative features, morphosyntactic patterns, and semantic distributions suggest that naive deployment of LLMs for social simulation may introduce systematic biases that compromise research validity. The observation that even fine-tuned models, while substantially improved, still exhibit detectable patterns in classification tasks suggests that the challenge extends beyond simple technical optimization to fundamental questions about the nature of human-like generation.

These findings align with broader concerns about the anthropomorphism of AI systems \cite{salles2020anthropomorphism} and highlight the necessity for validation protocols when deploying LLMs in social research contexts \cite{moller2024prompt}. The consistent performance hierarchy observed across all feature extraction approaches, with fine-tuned models showing highest alignment to human content, followed by prompt-based models, while the two synthetic approaches exhibit lowest mutual similarity, provides empirical evidence for the complexity of achieving authentic human simulation.

\subsection{Linguistic Authenticity and Model Limitations}
The systematic differences we observe across quantitative linguistics, morphosyntactic patterns, and semantic distributions point to inherent limitations in current language modeling approaches. Our analysis reveals that prompt-based models exhibit distinctive linguistic signatures, including more complex sentence structures (evidenced by coordinating and subordinating conjunction usage patterns), more topically diverse content, and significantly higher proportions of positive emotion classifications compared to human content.

Particularly concerning is the cluster-based similarity analysis, which shows the most pronounced differences between generation approaches. The substantial gaps between original \& prompt similarity and original \& fine-tuned similarity demonstrate that semantic distributional properties are particularly sensitive to generation method. These findings suggest that LLMs may be systematically biased toward producing "ideal" rather than authentic communication, potentially missing the natural variation, errors, and stylistic inconsistencies that characterize genuine human social media discourse \cite{thapa2025large}.

The cross-linguistic consistency of these patterns across English and German corpora strengthens the generalizability of our findings, indicating that the observed limitations are not language-specific artifacts but reflect fundamental characteristics of current language modeling approaches.

\subsection{Methodological Considerations for LLM Deployment}
The superior performance of fine-tuned models compared to prompt-based approaches across all similarity metrics provides strong evidence for the importance of domain adaptation in social media generation tasks. Fine-tuned models consistently achieve higher similarity scores with human content compared to prompt-based models. However, the persistence of detectable patterns even after fine-tuning, suggests that current adaptation techniques may be insufficient for achieving true linguistic authenticity \cite{munker2025don}.

The effectiveness of different encoding approaches in our validation task reveals important insights about the nature of synthetic content detection. The surprising performance of traditional tf-idf representations, particularly for prompt-based content detection, suggests that surface-level lexical patterns remain highly discriminative despite the sophistication of modern language models. The superior performance of hybrid approaches combining tf-idf, fastText embeddings, and extracted linguistic features demonstrates that multiple representational perspectives are necessary to capture the full spectrum of linguistic differences between human and synthetic content.

\section{Conclusion}
\label{sec:conclusion}

Our paper has examined a fundamental question about the viability of LLMs as human simulacra in computational social science: can current generation techniques produce social media content that reliably replicates authentic human linguistic behavior? Through systematic analysis of a novel history-conditioned dataset spanning English and German X content, we provide evidence-based answers to three interconnected research questions.

\subsection{Research Questions}

\paragraph{RQ1: Linguistic Pattern Detection} 
Our results demonstrate that LLM-generated social media posts exhibit systematic and detectable linguistic patterns across quantitative, morphological, and semantic dimensions. The similarity analysis reveals that, while individual synthetic posts may appear plausible, aggregate patterns consistently deviate from human norms. Most notably, prompt-based models show distinctive signatures in morphosyntactic complexity, with systematic differences in conjunction usage patterns indicating artificially complex sentence structures compared to human originals. Semantic analysis reveals systematic biases toward positive emotion classifications and increased topical diversity compared to authentic human content.

\paragraph{RQ2: Fine-tuning versus Prompt-based Approaches} 
Fine-tuned models consistently outperform prompt-based approaches across all similarity metrics, achieving substantially higher alignment with human content. However, even fine-tuned models remain distinguishable from human content in classification tasks, particularly through cluster-based similarity analysis where the most pronounced differences emerge. This finding confirms that training models with human data for concrete, well-defined tasks consistently outperforms general prompt-based usage approaches, aligning with findings from concurrent work demonstrating the limitations of generic prompting strategies \cite{munker2025don}.

\paragraph{RQ3: Machine Learning Detection Capability} 
Our validation task demonstrates reliable classification performance across multiple encoding approaches and feature combinations. The highest performing hybrid approach (tf-idf + fastText + extracted features) achieves macro $F1$ scores of $0.7301$ (German) and $0.6972$ (English), with particularly strong detection rates for prompt-based content ($F1$ > $0.8$ across multiple configurations). Surprisingly, traditional tf-idf representations prove remarkably effective, suggesting that surface-level lexical patterns remain highly discriminative despite advances in generation sophistication.

\subsection{Recommendations for Responsible LLM Deployment}
Based on our findings, we propose specific guidelines for the responsible deployment of LLMs in social applications:

\paragraph{Mandatory Validation Protocols}
Researchers employing LLMs for social simulation must implement comprehensive validation protocols that assess linguistic authenticity across multiple dimensions rather than relying on surface-level plausibility assessments. Our multi-dimensional evaluation framework provides a template for such validation, combining quantitative linguistics analysis, morphosyntactic profiling, semantic classification, and distributional similarity measures.

\paragraph{Domain-Specific Fine-tuning Requirements}
Our results confirm that fine-tuned models consistently outperform prompt-based approaches for social media generation tasks across all similarity metrics. However, fine-tuning alone proves insufficient to achieve complete linguistic authenticity, as evidenced by persistent detectability in classification tasks. This suggests that domain adaptation should be considered a minimum requirement rather than a sufficient solution.

\paragraph{Multi-dimensional Evaluation Standards}
The complementary nature of different linguistic analysis approaches in our study demonstrates that single-metric evaluation is insufficient for assessing generation quality. Researchers should adopt multi-layered evaluation frameworks that capture quantitative features, morphosyntactic patterns, semantic distributions, and embedding-based similarity measures simultaneously.

\subsection{Future Directions}
Our findings open several relevant directions for future research. First, investigating the temporal stability of linguistic signatures as generation techniques continue to evolve will be essential to understand the longevity of current detection methods and to develop robust evaluation frameworks. Second, examining domain transfer across different social media platforms beyond X will help establish the generalizability of these linguistic signature patterns across diverse communication contexts with varying discourse norms and constraints.

Third, exploring adversarial training approaches specifically designed to reduce detectability while maintaining content quality and authenticity represents a promising direction for improving generation fidelity. Such approaches could inform the development of more sophisticated LLMs that better capture the natural variation, errors, and stylistic inconsistencies characteristic of genuine human discourse on social media. Finally, developing more nuanced evaluation metrics that capture subtle aspects of human communication patterns beyond current similarity measures could provide deeper insights into the fundamental challenges of achieving truly human-like text generation.

The cross-linguistic consistency of our findings across English and German corpora suggests that these challenges transcend language-specific artifacts and reflect fundamental limitations in current language modeling approaches.

\section*{Limitations}
Our analysis focuses on X data collected during the first half of 2023, which may not generalize to other social media platforms or communication contexts with different discourse norms and constraints. The temporal dimension of our dataset may not capture evolving generation capabilities as LLM technology continues to advance rapidly. Additionally, our current framework focuses on English and German languages, and expanding the analysis to include morphologically richer languages, tonal languages, and non-European linguistic families would strengthen the cross-linguistic validity of these findings. Beyond these core limitation, we acknowledge several methodological.

\paragraph{Analysis Framework}
Our linguistic analysis framework, while comprehensive across quantitative, morphosyntactic, and semantic dimensions, does not capture complex discourse quality metrics such as argumentation coherence, irony detection, or cultural nuance recognition. The focus on individual post generation rather than multi-turn conversational dynamics limits our understanding of how synthetic content would perform in sustained social interactions and community discussions.

\paragraph{Validation Experiments}
Our detection validation experiments, while demonstrating reliable classification performance, are limited to the specific LLM architectures and fine-tuning approaches employed in this study. The rapid evolution of language models means that newer generation techniques may exhibit different linguistic signatures than those captured in our analysis. Additionally, our evaluation framework relies primarily on automated feature extraction and classification metrics, which may not capture subtle qualitative differences that human evaluators would detect.

\paragraph{Single Model Architecture}
Our results are based exclusively on Qwen3 8B, which represents only a single model architecture and size configuration. The observed linguistic patterns and detection accuracies may vary considerably across different model families, model sizes, quantization approaches within the same base model, and model versions. This architectural specificity limits the generalizability of our findings to the broader landscape of available LLMs.

\paragraph{Reply Prediction Task}
The history-conditioned reply prediction task relies on only three prior tweet-reply pairs as context, which may provide sparse predictive signal for capturing individual user behavior patterns and writing styles. This limited historical context may not fully represent the complexity and variation present in users' broader communication patterns, potentially affecting both the fine-tuning quality and the authenticity of generated content.

\paragraph{German vs. English}
The German and English datasets differ substantially in their collection contexts, temporal distribution, and underlying discourse characteristics. These systematic differences make direct cross-linguistic performance comparisons not recommended, as observed variations may reflect dataset-specific properties rather than fundamental linguistic or modeling differences. Each language corpus should be interpreted within its own context rather than as directly comparable benchmarks.

\section*{Ethical Considerations}
As is typical for AI methods, the modeling approach presented in this paper is a dual-use technology. While behavior-based user modeling and synthetic content generation are primarily intended for computational social science research and platform safety applications, the findings can also be used to develop more sophisticated manipulation techniques or improve the convincingness of synthetic social media content for malicious purposes.

\paragraph{Privacy and Consent Considerations}
A significant ethical concern in our study involves the use of real user data from X to train models that replicate individual behavior patterns. While our dataset consists of publicly available posts from political discourse and replies from regular users, the individuals whose data we used did not provide explicit informed consent for their communication patterns to be learned and replicated by generative models. This raises important questions about digital privacy rights, even when dealing with publicly posted content.

\paragraph{Potential for Misuse}
The detection methodologies developed in this work, while intended to improve synthetic content identification, could potentially be used adversarially to develop more sophisticated generation techniques that evade detection. The detailed analysis of linguistic signatures across quantitative, morphosyntactic, and semantic dimensions provides a road-map for improving synthetic content quality, which could enhance both legitimate applications and malicious use cases.

\paragraph{Broader Implications}
The development of increasingly sophisticated user modeling and synthetic content generation capabilities raises broader questions about the boundaries of acceptable research practices in computational social science. As these technologies advance, the research community must carefully balance the scientific value of realistic behavioral simulation against the privacy rights and dignity of individuals whose data enables such research, while considering the potential societal impacts of increasingly convincing synthetic social media content.

\section*{Acknowledgments}
We thank Simon Werner and Christoph Hau for our constructive discussions. This work is supported by TWON (project number 101095095), a research project funded by the European Union under the Horizon framework (HORIZON-CL2-2022-DEMOCRACY-01-07).


\section*{Bibliographical References}
\label{sec:reference}
\bibliographystyle{_external/lrec_natbib.bst}
\bibliography{main}

\end{document}